\documentclass[11pt,letterpaper]{llncs}
\usepackage{amsmath}
\usepackage{amssymb}
\usepackage{graphicx}
\usepackage{marvosym}
\usepackage{url}
\usepackage{fancyhdr}
\usepackage{hyperref}

\usepackage{geometry}
\newcommand{\keywords}[1]{\par\addvspace\baselineskip
\noindent\keywordname\enspace\ignorespaces#1}

\fancypagestyle{alim}{\fancyhf{}\fancyfoot[L]{\scriptsize{Dhinaharan Nagamalai et al. (Eds): CSEIT, WiMoNe, NCS, CIoT, CMLA, DMSE, NLPD - 2020 \\  
pp. 01-16, 2020. CS \& IT - CSCP 2020 ~ ~~~~~~~~~~~~~~~~~~~~~~~~~~~~~~~~~~~~~~~~~~~~~~~~~ \href{https://aircconline.com/csit/papers/vol10/csit101101.pdf}{DOI: 10.5121/csit.2020.101101}        
}}}

\usepackage{etoolbox} 
\makeatletter 
\patchcmd{\ps@headings}{{\slshape\rightmark}\hfil\thepage}{\thepage\hfil}{}{}
\begin{document}

	
	\title{\LARGE{Evaluating the impact of different  types  of crossover and selection methods on the convergence of 0/1 Knapsack using Genetic Algorithm}}

	\author{\large{Waleed Bin Owais \and Iyad W. J. Alkhazendar  \and Dr.Mohammad Saleh  }}
	
	\institute{Department of Computer Science and Engineering,\\Qatar University,Doha}

	\maketitle

	\begin{abstract}
		Genetic Algorithm is an evolutionary algorithm and a metaheuristic that was introduced to overcome the failure of gradient based method in solving the optimization and search problems. The purpose of this paper is to evaluate the impact on the convergence of Genetic Algorithm vis-a`-vis 0/1 knapsack. By keeping the number of generations and the initial population fixed, different crossover methods like one point crossover and two-point crossover were evaluated and juxtaposed with each other. In addition to this, the impact of different selection methods like rank-selection, roulette wheel and tournament selection were evaluated and compared. Our results indicate that convergence rate of combination of one point crossover with tournament selection, with respect to 0/1 knapsack problem that we considered, is the highest and thereby most efficient in solving 0/1 knapsack.
	\end{abstract}

	\section{Introduction}
	A genetic algorithm can be defined as a search heuristic algorithm that is motivated by Darwin’s theory of natural evolution. As mentioned, this is motivated by Darwinian Natural Section and then applies them to soft computing. In Darwinian Natural Section there are three key principles that need to be in place for evolution.[13]
	\begin{itemize}
		\item 	Hereditary: A procedure by which children obtain the characteristics of their parents.
		\item 	Variation: Their should be an element of variety in a population, that is, it should not be homogeneous throughout.
		\item Selection: A mechanism by which some members of the population have the opportunity to be the parents and pass down their genetic information and some do not. Also known as the survival of the fittest [1].
		There are five phases associated with a genetic algorithm
		1.	Initial population 2. Fitness function 3. Selection 4. Crossover 5. Mutation [2]
\thispagestyle{alim}
	\end{itemize}

	\section{DEFINITIONS}
	The paper in later section's  uses some of the technical keywords that are related to the Genetic Algorithm so it is imperative to define such technical keywords.
	\begin{itemize}
		\item 	Initial Population:
		As illustrated in the Figure 1, in a genetic algorithm, the set   of genes of an individual is characterized using a string of 1s and 0s is used. This can be referred to as coding. Genes can  be defined as a single element . The genes then join together into a string to form a Chromosome (solution)[18].
		\item 	Fitness Function:
		The fitness function can be defined as a function that helps    in determination of how fit an individual is and gives an inference about its ability to compete with other individuals. The fitness function usually assigns a fitness score to each individual [14].
		\item 	Crossover:
		Crossover is the most significant stage in a genetic algorithm. Two chromosomes are mated by choosing a point of crossover. The crossover in Genetic Algorithm creates new generation  as the same is done in the process of natural mutation.. The resulting chromosomes are known as offspring’s [3]	\\
		\item 	Mutation:
		In some of the new offspring made, some of the genes are subjected to a mutation. In soft computing the purpose of mutation is to ensure that the solution is not stuck at local optima and the solution explores the entire search  space in the pursuit of finding the global maxima or global minima [17].\\
	\end{itemize}
	\begin{figure}
		\includegraphics{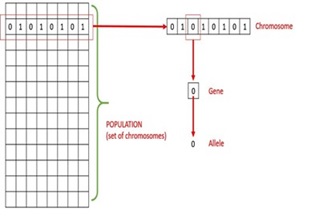}
		\caption{Various Parameters}
		\label{fig:1}       
	\end{figure}

	This paper will compare the result of various crossover techniques and selection methods in 0-1 knapsack. The knapsack problem deals with the idea of filling the knapsack with different items to maximize the profit without exceeding the net capacity of the knapsack. In other words, it is maximizing the profit while minimizing the cost. We will use a combination of most widely used crossover methods and selection methods and observe the results by doing a one to many mapping between the crossover methods and selection techniques.

	\section{BACKGROUND AND RELATED WORK}

	In the field of combinatorial optimization, knapsack problem can be defined as the process of finding an optimal solution from a finite set. The aim is to maximize the profit by including the items(each having a given weight and value) in the knapsack without exceeding the total capacity of the knapsack. 0/1 knapsack means that there is an additional constraint of either selecting the item in its entirety(1) or not selecting the item at all(0).\\
	This can be explained as follows:\\
	A set of finite items from 1 to n, each having a value of v    and weight of w, and x being the number of copies of each item, whilst the total weight of knapsack being W, then based on 0/1 knapsack we have:\\
	\begin{equation}
	\sum_{i=1}^{n}wi*xi\\
	\end{equation}

	Subjected to the following constraints:\\
	\begin{equation}
	\sum_{i=1}^{n}wi*xi\leq W,xi\in 0,1\\
	\end{equation}
	Knapsack problem has found a lot of applications in real world for example in making decisions related to investment banking, selection of project, and vote trading problem. In the research related to comparing the results of 0/1knapsack much of the research has been done in comparing the results of Genetic Algorithm with Greedy Approach , Branch and Bound and dynamic algorithm [3]-[5]. In another comparative study titled “ Comparative study of meta-heuristic algorithms using Knapsack Problem” [6] the authors have compared various meta-heuristic techniques to solve knapsack problem. Also [7] used chaotic crossover operator on Genetic Algorithm which produced improved results. In the research by [8] job scheduling problem(an application of 0/1 knapsack) was solved using Genetic Algorithm performance of various crossover techniques was presented. Similarly in [9] the research 
	evaluated the impact of various crossover techniques on a web classifier. In [10]  the authors have compared six crossover techniques and evaluated their performance on 0/1 knapsack. The experiments that followed, two point crossover showed the best results. In [11] and [12] the authors have evaluated the performance of various selection techniques. In [15] the authors evaluated various algorithmic techniques used in optimization of 0/1 knapsack.  \\
	To the best of the knowledge of author's, none of the research hitherto has combined various selection and crossover methods and evaluated their performance on convergence of 0/1 knapsack using the Genetic Algorithm. 
	
	\section{	WORKING OF GENETIC ALGORITHM}
	This section describes the working of Genetic Algorithm \\
	Step 1: Generating the initial population randomly.\\
	Step 2: Calculating the ﬁtness of the population. \\
	Step 3: Selecting the ﬁttest individuals based on ﬁtness.\\
	Step 4: Producing offspring’s by crossover of selected chromosomes.\\ 
	Step 5: Applying mutation.\\
	Step 6: Go back to step 2 until termination condition is satisﬁed. 
	\begin{figure}
		\includegraphics{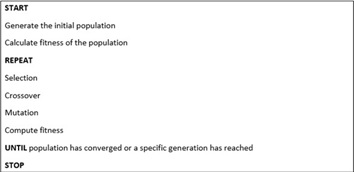}
		\caption{Psuedocode of Genetc Algorithm}
		\label{fig:2}       
	\end{figure}
	
	\section{ CROSSOVER METHODS }
	As described crossover, also known as recombination is the most imperative process in the stages of Genetic Algorithm. It is in this phase that two parents exchange genetic information by choosing a single or multi point of crossover. Crossover operators divides a pair of selected chromosomes into two or more parts. After that the combination of chromosomes takes place to produce a new offspring (child). There are two types of crossover’s that we have used: 
	\begin{figure}
		\includegraphics{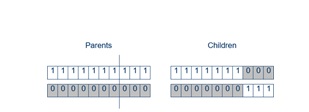}
		\caption{Diagrammatic representation of Single Point Crossover}
		\label{fig:4}       
	\end{figure}
	\subsection{ Single Point Crossover }
	In a single point crossover only one point is designated as a crossover location.After a random point is chosen, the parents slip at the crossover point and offspring’s are created by exchanging tails. In the 
	\begin{figure}
		\includegraphics{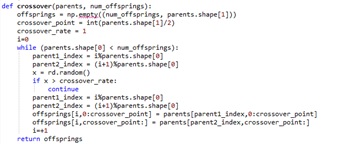}
		\caption{Code Snippet of Single Point Crossover}
		\label{fig:3}       
	\end{figure}
	
	Figure 4 as can be seen the 8th point is chosen as crossover point and the bits to the right side of the crossover point (111 and 000) are exchanged inter alia.
	
	\subsection{ Two Point Crossover }
	In the two point crossover, the there are two points wherein the exchange of information takes place. The information between these two points is exchanged between parents to form offspring’s.
	
	\begin{figure}
		\includegraphics{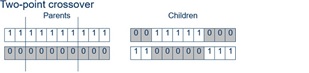}
		\caption{Diagrammatic representation of Two Point Crossover}
		\label{fig:5}       
	\end{figure}
	As can be seen in Figure 5, the 2nd point and the 7th point are designated as the two points for the crossover. The genes between them are swapped (11111 and 00000) to create two new offspring’s.  Figure 6 is a snapshot of coding scheme used in Two Point crossover.
	\begin{figure}
		\includegraphics[width=10cm,height=8cm,keepaspectratio]{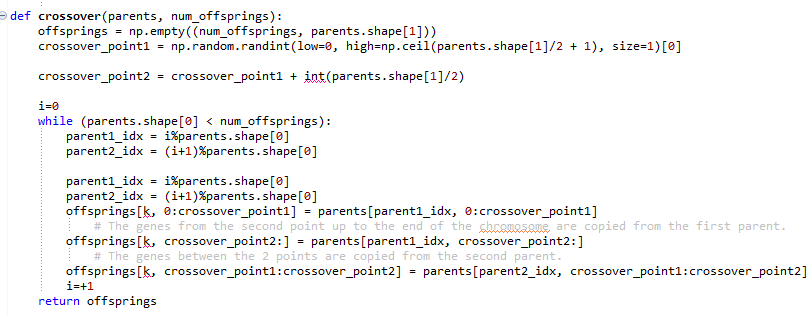}
		\caption{Code Snippet of Two Point Crossover}
		\label{fig:6}       
	\end{figure}
	\section{  SELECTION METHODS  }
	In each generation, based on certain pre-determined criteria, only some of the chromosomes of the population are chosen to take part in the mating process, which is the crossover and mutation. This ﬁltering of the population based on the ﬁtness function is known as selection. The purpose of selection is to choose only those chromosomes who satisfy a speciﬁc criteria with regards to the ﬁtness of the chromosome.\\ Such ﬁltering of the population ensures that only healthy individuals are promoted to the next generation and the unhealthy ones (who do not satisfy a criteria) are left behind. \\For instance, consider the solution set of a 0/1 knapsack. Any subset of population that has Formula(1), will not be considered for selection, because it doesn’t satisfy the ﬁtness function, because the weight exceeds the total weight of the knapsack. Fitness function is the criteria that is used to ﬁlter chromosomes in the selection step. Not only does the selection do the ﬁltering of the ﬁt chromosomes from the unﬁt chromosomes,it also helps in arranging the chromosomes based on their ﬁtness. \\For instance in the max one problem, the chromosome’s are arranged in a hierarchy with the most ﬁt( the chromosome having maximum 1’s )at the top. It should be noted that in the process of selection, ﬁltering based on the ﬁtness function doesn’t always take place, as can be seen in the case of maxone problem, because all the chromosomes satisfy the ﬁtness function. In such cases the ordering of chromosome’s based on their ﬁtness takes place. There are a lot of selection methods that are used in the Genetic Algorithm. We have used the following : 
	
	\subsection{  Rank Selection  }
	In the method of rank selection, after assignation of ﬁtness to each individual, they are arranged in decreasing order of rank, in other words, the most ﬁt individual gets rank one and so on. After each individual is assigned a rank, the chromosomes have a chance to get selected.The probability of an individual getting selected is given by the formula: \\
	\begin{equation}
	\rho (i)=rank(i)/n*(n-1)
	\end{equation}
	
	where p is the probability of individual i and n is the total
	number of individuals competing.\\
	For instance if chromosome’s 1 through 5 have a fitness of 37,6,36,30, and 28 respectively, then on the basis of rank selection, the individuals will be ranked as 1,3,4,5,2 based on their fitness. Chromosome 1 has the first rank and chromosome 2 has the last rank.
	\begin{figure}
		\includegraphics{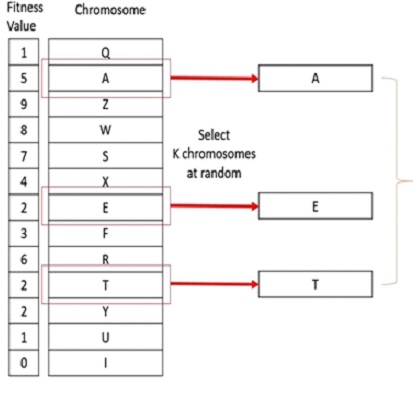}
		\caption{Diagrammatic representation of Tournament Selection}
		\label{fig:7}       
	\end{figure}
	\subsection{  Roulette Wheel Selection }
	Roulette wheel selection, also known as Fitness Proportionate Selection, is another widely used selection method in Genetic Algorithm. After each individual is assigned the fitness score via fitness function, the roulette wheel determines the selection. The higher the fitness of an individual, the higher the chances will be  to  get  selected. The process in roulette selection does a linear search through  a roulette wheel where each individual gets a share in the roulette wheel. That is, higher  the  fitness,  higher  will  be  the share in the wheel, and thereby higher chance of to be selected when the wheel spins. Weaker individuals, having  less share in the wheel, have very less probability of getting selected.\\
	The probability of an individual to get selected via roulette wheel selection is given by the formula:\\
	\begin{equation}
	\rho (i)=f(i)/\sum_{j=1}^{n}f(j)
	\end{equation} 
	
	where p is the probability of individual i, f is the fitness      of individual i and f(j) is the total fitness of the population.    In regards to the simplicity and easiness of implementation, the roulette wheel selection is the most preferred method of selection.
	
	\subsection{ Tournament Selection }
	
	In this selection method,shown in Figure 7, the  individuals contend against each other. The one with the highest fitness apparently wins the tournament  and  is  selected  for the subsequent generation. Weak individuals(one having low fitness) have less chances to be selected. 
	The number of the chromosomes that contend against each other is termed as tournament size. The default tournament size is 3. It should   be noted that in tournament selection,  every chromosome is given an equal chance to compete.
	
	\section{ Results }
	
	In the implementation of various crossover techniques and selection methods, we used Python 3.7.4, in the PyDev module and implemented Genetic Algorithm on the following:\\
	0/1 knapsack which has the following characteristics: \\
	Weight = [2, 3, 6, 7, 5, 9, 4, 5, 2, 3, 4, 1, 7, 8, 4, 5, 3]\\
	Value = [6, 5, 8, 9, 6, 7, 3, 7, 4, 2, 5, 8, 3, 1, 5, 2, 8]\\
	Knapsack threshold = 29 \\
	There are a total of 17 items that can be chosen as 1 or left over as 0, and the respective weights and values are given. The aim is to maximize the proﬁt of the knapsack without breaking the knapsack, that is the net weight should be less than or equal to Knapsack threshold. The genetic algorithm ran 20 times for each scenario that will be discussed fore with. The following parameters were constant throughout the experiment.\\ Number of generations= 50\\ Solutions per population =8(No. of chromosome’s within each population )\\ Mutation Rate=0.4\\ Type of mutation=Bit ﬂip mutation\\ Crossover Rate=0.8 \\The Genetic Algorithm was applied in the following scenarios.
	Although attributed to the 'No Free Lunch Theorem' [16], there is no best crossover or selection technique but the authors chose the following selection methods and crossover techniques as they have been most widely used in research done thus far and shown interesting results.
	\subsection{ Scenario 1 }
	As shown in figure 8\begin{figure}
		\includegraphics{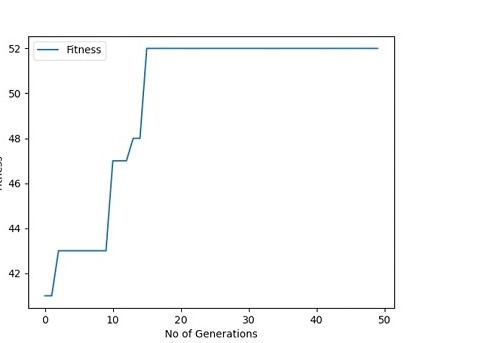}
		\caption{Combination of OnePoint Crossover with Rank Selection}
		\label{fig:8}       
	\end{figure} 
	,in this scenario the type of the crossover is one point whilst the selection method is rank selection.
	The remaining parameters deﬁned above remain constant. The results show a spike towards the optimal solution from the 15th generation onwards. The optima was stuck at 43 for about 9 generations. 
	\subsection{ Scenario 2 }
	As shown in figure 9,\begin{figure}
		\includegraphics{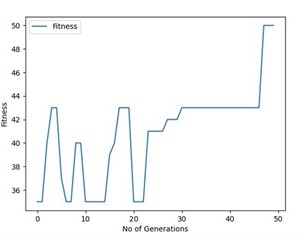}
		\caption{Combination of OnePoint Crossover with Roulette Wheel Selection}
		\label{fig:9} 
	\end{figure}
	In this scenario the type of the crossover is one point whilst the selection method is roulette wheel selection. The remaining parameters defined above remain constant.\\
	
	As is evident from the figure, this combination is quite slow as it does not converge to optimal solution in the ﬁxed 50 generations. There can also be seen a trend of ﬁtness increasing and decreasing till the 20th generation, and then it gets stuck at local optima of 43 for about 25 generations. The results indicate that the combination of One point crossover with roulette wheel selection is slow to converge to optimal solution.
	
	\subsection{ Scenario 3 }
	As shown in ﬁgure 10,in this scenario, the type of the crossover is one point whilst the selection method is tournament selection. The remaining parameters deﬁned above remain constant.\\
	\begin{figure}
		\includegraphics{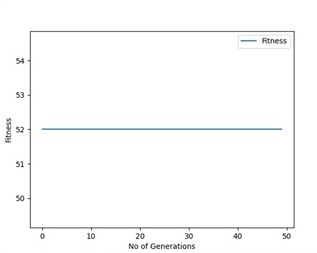}
		\caption{Combination of OnePoint Crossover with Tournament Selection}
		\label{fig:10}       
	\end{figure} 
	The results indicate that the combination of one point crossover with tournament selection converges as quickly as 1st generation and remains constant till the end of the 50th generation.
	
	\subsection{ Scenario 4 }
	
	As shown in ﬁgure 11,in this scenario we have combined two point crossover with the rank selection method.The remaining parameters deﬁned above remain constant.\\
	As is evident from the ﬁgure the solution is stuck at local optima of 40 for about 8 generations, where it shows a steep increase \begin{figure}
		\includegraphics{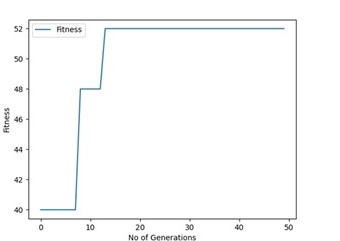}
		\caption{Combination of TwoPoint Crossover with Rank Selection}
		\label{fig:11}       
	\end{figure} to 48,and after getting stuck at 48,it gives the optimal solution from the 15th generation. No much variation is seen, the optima changes just two values before converging at global optima.
	
	\subsection{ Scenario 5}
	
	As shown in ﬁgure 12,in this we have combined two point crossover with roulette wheel selection method. The remaining parameters deﬁned above remain constant. The results do not show a healthy trend, ﬁrst it doesn’t converge to the optimal solution of 52 in the ﬁxed 50 generations, and,\begin{figure}
		\includegraphics{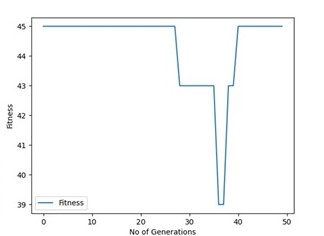}
		\caption{Combination of TwoPoint Crossover with Roulette Selection}
		\label{fig:12}       
	\end{figure}  as is evident from the ﬁgure after getting stuck at local optima of 45 there is a decrease in ﬁtness from 25th generation onwards where the ﬁtness decreases to 39 and again increases to 45 subsequently. Results indicate that convergence of this combination is very slow. 
	\subsection{ Scenario 6}
	
	As shown in ﬁgure 13,\begin{figure}
		\includegraphics{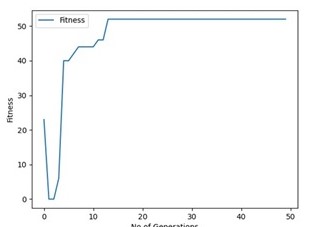}
		\caption{Combination of TwoPoint Crossover with Tournament Selection}
		\label{fig:123}       
	\end{figure} in this we have combined two point crossover with tournament selection method. The remaining parameters deﬁned above remain constant. The results indicate that optimal solution converged to 52 at the beginning of the 20th generation. Initially the optimum showed a downward trend, where the ﬁtness even fell to 0. 
	
	\section{ Discussion}

	In the six combinations that we tested, all the combinations converged except for the combination of two point crossover method with the roulette wheel selection method. Although this combination will converge too if we increase the number of generations. Since we used 50 generations as baseline in all of the experiments, we can say that the convergence of this combination is slow. Also the combination of one point crossover with roulette wheel selection doesn’t converge to optimal solution in 50 generations, but it is faster than the combination of two point crossover method with the roulette wheel selection method, as it converges to 50 as against to 45 of latter. \\
	On the contrary, the results show that the convergence rate of combination of one point crossover with tournament selection converges as quickly as 1st generation and remains constant until the end of the 50th generation. The optimal solution of 52 is achieved in the ﬁrst-generation itself. In the six combinations, the combination of one point crossover with tournament selection remains the fastest with respect to the 0/1 knapsack problem that we considered. The other combination methods converge to the optimal solution of 52, but we conclude that convergence rate of combination of one point crossover with tournament selection, with respect to 0/1 knapsack problem that we considered, is the highest and thereby most efficient in solving 0/1 knapsack.\\
	In future endeavors we would like to test this on a randomly chosen sample space of weight and values and adjust the knapsack threshold accordingly.It would also be interesting to combine Genetic Algorithm with other meta heuristic algorithms and gauge the impact on the convergence of 0/1 knapsack thereof and test whether it has any substantial impact on the reduction of number of iterations and time of convergence. Another possible future work would be evaluating the impact of various types of mutation vis-a-vis the scenarios aforementioned.\\
	The properties of the computer used in experimentation are Intel(R) Core(TM) i7-4600U CPU @ 2.10 GHz 2.69 GHz, and 8GB RAM with x64 based processor.The algorithm is written in Python 3.7 in the PyDev module of Eclipse.
	\section{Conclusion}
	
	The knapsack problems have a wide variety of applications in real world like cargo loading, budgeting, project management et.al. In our paper we combined different types of crossover methods with different selection techniques and evaluated their rate of convergence. We concluded that the combination of one point crossover with tournament selection is the most efficient. We also have discussed some of the future directions that the authors would like to work on.
	
	\vspace{2cm}
	
	\section*{Authors}
	\noindent {\bf Waleed Bin Owais} received his Bachelor's of Technology from BGSBU, India, and is currently pursuing his MS at Qatar University, Doha. His research interests include Virtual Reality, Empathy, Nuerofeedback and Algorithms.\\
	
	\noindent {\bf Iyad W. J. Alkhazendar }received Bachelor's degree in  computer science from Al-Azhar University, Palestine and is currently pursuing his MS at Qatar University, Doha. His research interests include Network security, communication, Industrial IOT, SCADA , and Algorithms. \\
	
	\noindent {\bf Dr. Mohammad Saleh } received his PhD, Computer Science(Computer Networking) from University Putra Malaysia (UPM) and is currently Associate Professor in Department of Computer Science and Engineering, College of Engineering, Qatar University. His research interests include Simulations and modeling, OO Metrics.  \\
	\\    
	
	\noindent{\textcopyright}2020 By AIRCC Publishing Corporation. This article is published under the Creative Commons Attribution (CC BY) license.
\end{document}